# System Report for CCL24-Eval Task 7: Multi-Error Modeling and Fluency-Targeted Pre-training for Chinese Essay Evaluation

**Jingshen Zhang**[†], **Xiangyu Yang**[†], **Xinkai Su, Xinglu Chen, Tianyou Huang, Xinying Qiu**[‡]

School of Information Science and Technology,

Guangdong University of Foreign Studies, Guangzhou, China

{20211003207,20221050039,20210602169,20221003046,20231010040}@gdufs.edu.cn

xy.qiu@foxmail.com

## Abstract

This system report presents our approaches and results for the Chinese Essay Fluency Evaluation (CEFE) task at CCL-2024. For Track 1, we optimized predictions for challenging fine-grained error types using binary classification models and trained coarse-grained models on the Chinese Learner 4W corpus. In Track 2, we enhanced performance by constructing a pseudo-dataset with multiple error types per sentence. For Track 3, where we achieved first place, we generated fluency-rated pseudo-data via back-translation for pre-training and used an NSP-based strategy with Symmetric Cross Entropy loss to capture context and mitigate long dependencies. Our methods effectively address key challenges in Chinese Essay Fluency Evaluation.

**Keywords:** Chinese Easy Fluency Evaluation , Error Sentence Type Recognition , Error Sentence Correction

## 1 Introduction

With the growing integration of smart education and deep learning technologies, automated text evaluation systems have become increasingly critical. These systems aim to accurately and efficiently assess students' compositions, reduce teachers' workload, and provide instant feedback for error correction and writing improvement. The China National Conference on Computational Linguistics (CCL-2024) has presented the Chinese Essay Fluency Evaluation (CEFE) as a public assessment task. This task focuses on three primary text evaluation strategies: error sentence type recognition, error sentence correction, and essay fluency evaluation, offering an in-depth research direction for the field of automatic text evaluation.

This system report provides an overview of our work on the CEFE evaluation task, highlighting the different strategies employed for each track:

- For Track 1, Error Sentence Type Recognition, we analyzed two fine-grained errors, utilized a binary classification model for prediction optimization, compared and selected training corpora, and trained a coarse-grained model based on the Chinese Learner 4W corpus.

- For Track 2, Error Sentence Correction, we adopted a strategy that involved constructing a pseudo-dataset containing sentences with multiple error types to enhance model performance.

- For Track 3, Essay Fluency Evaluation, we achieved first place by employing back-translation techniques to construct pseudo-data with triple-labeled fluency ratings for pre-training and adapting an NSP-based strategy to effectively utilize contextual information and avoid long sequence dependencies.

The remainder of this report is structured as follows: Section 2 presents related research in the field of Chinese composition fluency evaluation. Sections 3, 4, and 5 detail our methodologies, experiments, and results for Tracks 1, 2, and 3, respectively. Finally, Section 6 concludes the report with an analysis of our findings, discusses the limitations of our work, and potential future research directions. Our code and data are available at https://github.com/astro-jon/ccl2024-coherence.

## 2 Related Research

The field of Chinese composition fluency evaluation has gained significant attention from researchers in recent years, with the three different track directions involved in this review being popular topics for related research.

### 2.1 Error Sentence Type Recognition

Error sentence type recognition has been a focus of many studies. Zhang et al., (2020) combined Graph Convolutional Networks (GCN) and Transformers for Chinese grammatical error detection, leveraging the strengths of both architectures to improve performance. Wang et al. (2023) proposed a multi-granularity approach for Chinese grammar error detection and correction, utilizing character-level, word-level, and sentence-level information to enhance the model's ability to identify and correct various types of errors.

### 2.2 Error Sentence Correction

Error sentence correction has also received significant attention in recent research. Li et al. (2022) proposed a Sequence-to-Action (S2A) module that combines source and target sentences as inputs to automatically generate token-level action sequences for predicting editing operations, effectively integrating the advantages of sequence-to-sequence (seq2seq) models and sequence-tagging models to mitigate the overcorrection problem and improve the performance of the syntactic error correction task. Wu and Wu (2022) introduced a new framework for Chinese grammatical error correction that addresses both spelling and grammar errors, utilizing a two-stage approach that first corrects spelling errors and then focuses on grammatical error correction. Zhou et al. (2023) proposed decoding interventions to improve seq2seq grammatical error correction models, focusing on enhancing the decoding process to generate more accurate and fluent corrections.

### 2.3 Essay Fluency Evaluation

In addition to error correction, the flow of the text is an equally critical factor in measuring the quality of the text. Mesgar and Strube (2018) proposed a neural local coherence model for text quality assessment that captures the flow of semantic connections between neighboring sentences based on the most similar semantic states and encodes the pattern of changes in text-perceived coherence. Qiu et al. (2022) explored the potential of coherence and syntactic features in neural models for automatic essay scoring, combining syntactic feature dense embedding with the BERT model and investigating the joint model of coherence, syntactic information, and semantic embedding. Sheng et al. (2024) proposed a novel non-referential coherence measure called BB Score, which is based on Brownian Bridge Theory and evaluates text coherence by measuring the ordered and coherent interactions between sentences.

## 3 Track 1: Error Sentence Type Recognition

### 3.1 Methodology

Our methodology for Track 1 employed a hierarchical approach as illustrated in Figure 1:

1. **Token-level error identification:** The approach starts by identifying errors at the token level rather than the sentence level. This step covers various types of errors, including

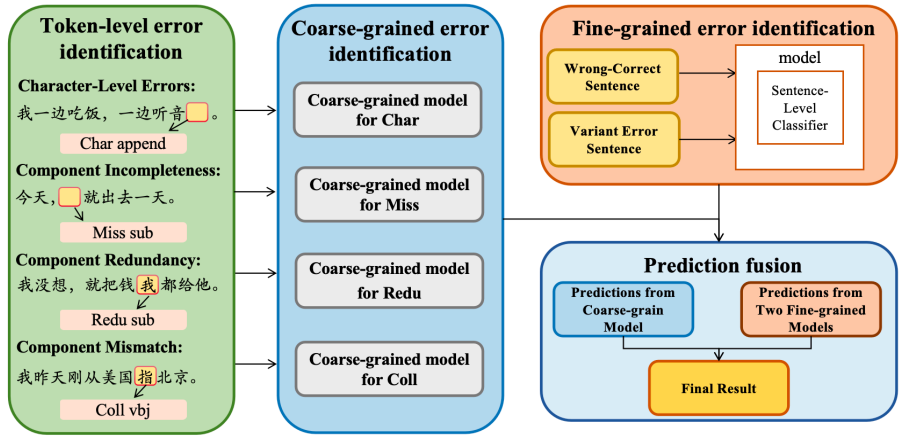

Figure 1: Methodology for Track1

Character-Level Errors (such as missing or incorrect characters), Component Incompleteness (e.g., missing subject), Component Redundancy (e.g., redundant subject), and Component Mismatch (e.g., verb-object mismatch).

2. **Coarse-grained error modeling**: Four separate coarse-grained error models are constructed to handle different error types: a model for Character-Level Errors (Char), a model for Component Incompleteness (Miss), a model for Component Redundancy (Redu), and a model for Component Mismatch (Coll).

3. **Fine-grained error categorization**: This step involves further refining the predictions from the coarse-grained models by categorizing and identifying specific fine-grained error types within each broad category. Two approaches are employed for this purpose:

   - **Wrong-Correct Sentence**: Sentences with errors are matched with their corrected counterparts, and the model learns to distinguish between erroneous and correct sentences.
   - **Variant Error Sentence**: Sentences containing specific error types (e.g., misordering) are paired with sentences containing other error types, and the model learns to differentiate between them.

4. **Prediction fusion**: The predictions from the four coarse-grained models and the two fine-grained models (Wrong-Correct Sentence and Variant Error Sentence) are combined to generate the final error identification results. This fusion step ensures that the insights from all the models are integrated to produce the most accurate and comprehensive error analysis.

### 3.2 Experiment and Results

We used the CSED (8,682 sentences) (Sun et al., 2023) and Chinese Learner 4W (39,989 sentences) (Lu et al., 2020) corpus for pseudo-data construction due to limited official data without token-level labels. For training our coarse-grained models, we trained each model on the respective datasets mentioned above instead of merging them. Subsequently, we selected the better one for each coarse-grained model.

The coarse-grained models were trained on the combined corpora using the chinese-electra-180g-base-discriminator model (Cui et al., 2021), with a maximum length of 512, 30 epochs, batch size 32, and learning rate 2e-5. We compared uniform (25% each) and full corpus distribution strategies.

For the fine-grained binary classification models, we used chinese-roberta-wwm with maximum length 512, 30 epochs, batch size 2, and learning rate 1e-5. Precision, recall, and micro F1 evaluated performance.

We successfully trained four coarse-grained models on the 4W corpus using non-repetitive pseudo-data construction. For the challenging misordering and redundancy error types, we trained fine-grained models on sentence pairs contrasting the target error with others using the public corpus. This approach achieved our best score of 36.47.

## 4 Track 2: Error Sentence Correction

### 4.1 Methodology

For Track 2, we observed that the original training set contained sentences with multiple error types, whereas previous pseudo-data construction methods from Wang et al. (2023) introduced only one error type per sentence. To better match the original data, we proposed constructing a pseudo-dataset containing sentences with varying numbers of error types with the following steps:

- Apply Wang et al.'s method to introduce single error types into correct sentences.

- Randomly select 1/5 of those single-error sentences.

- From the selected sentences in Step 2, randomly select another 1/5 and introduce a second error type.

- Repeat Step 3, selecting 1/5 from the previous iteration, to create sentences with up to four error types.

The proportion of constructed data can be estimated using the following formula:

$$Percent_i = C_3^{i-1}(1-p)^{4-i}p^{i-1} \tag{1}$$

where $i$ indicates the number of error types, and $p$ represents the selection parameter, which is set to 1/5 in this context to match the distribution of the original dataset.

Thus we created a diverse pseudo-dataset with sentences containing varying numbers of error types, better reflecting real-world erroneous data. We then trained the real_learner_bart_CGEC encoder-decoder model (Zhang et al., 2023) on this multi-error pseudo-dataset to enhance its ability to correct sentences with numerous errors.

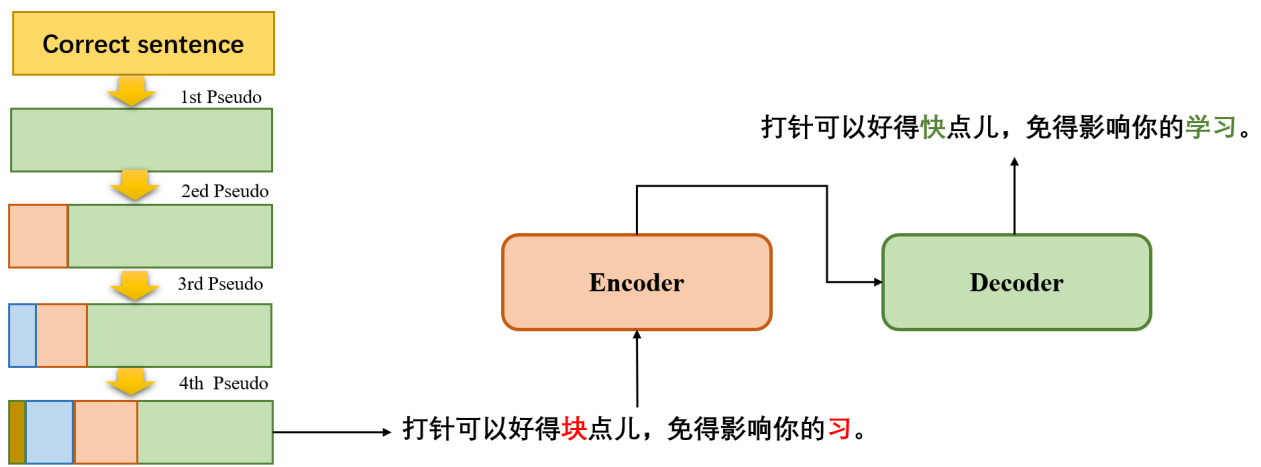

Figure 2: Methodology for Track 2

## 4.2 Experiment and Results

We used the Chinese Learner 4W Corpus (Lu et al., 2020) for our constructed corpus. We utilized the constructed pseudo-dataset containing numerous errors to train the Bart-base model and subsequently evaluated its performance on the validation set. The evaluation metrics were based on the calculations provided by the official Track 2 guidelines. The best result on the test set for Track 2 was achieved using the real_learner_bart_CGEC model proposed by Zhang et al. (2023), obtaining a final score of 41.09.

# 5 Track 3: Essay Fluency Evaluation

## 5.1 Methodology

For Track 3, we addressed two key challenges: limited training data and modeling document-level inputs. To augment the scarce labeled data, inspired by Lu et al., (2021), we employed back-translation to construct pseudo-data with triple fluency ratings, as shown in Figure 3(a). Essays back-translated using a resource-rich language, $lang_{rich}$, were labeled as moderately fluent, while those using a resource-poor language, $lang_{limit}$, were labeled as failing fluency. The original essays acted as excellent fluency examples. This back-translated corpus was used for pre-training. And we select English as $lang_{rich}$ and Japanese as $lang_{limit}$ for English has more translation training corpus than Japanese.

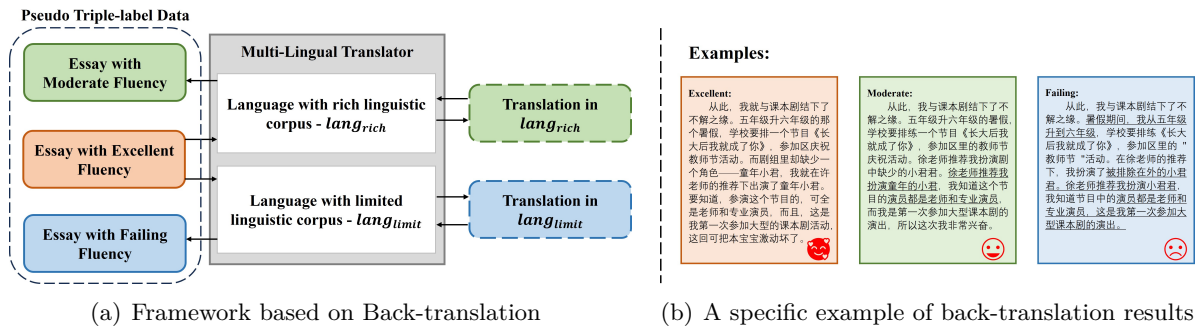

(a) Framework based on Back-translation  (b) A specific example of back-translation results

Figure 3: (a) Framework based on Back-translation; (b) A specific example of back-translation results

To capture contextual information while avoiding long sequence issues, inspired by Qiu et al. (2022), we adapted an NSP-based training strategy illustrated in Figure 4(c). Instead of inputting the entire essay (Figure 4a) or individual sentences (Figure 4b), we input pairs of neighboring sentences joined by [SEP] tokens. An average aggregation function combined the sentence pair predictions into a final essay fluency score.

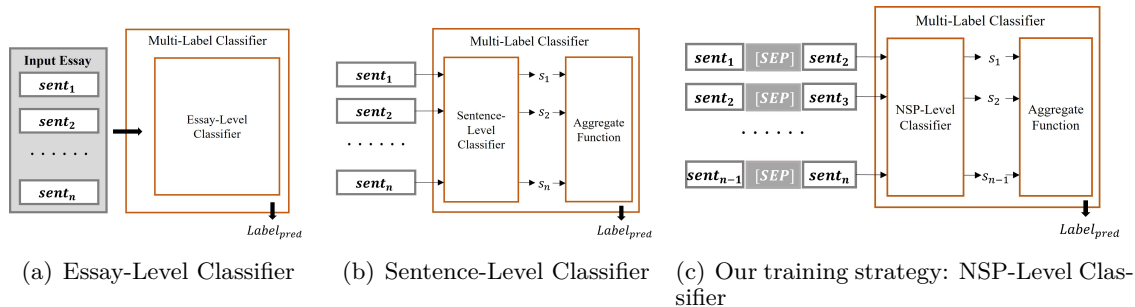

(a) Essay-Level Classifier  (b) Sentence-Level Classifier  (c) Our training strategy: NSP-Level Classifier

Figure 4: Comparison of different training patterns: (a) Essay-Level Classifier, (b) Sentence-Level Classifier, (c) Our training strategy: NSP-Level Classifier

To provide robustness to potential labeling noise from the pseudo-data, we optimized the Symmetric Cross Entropy (SCE) loss (Wang, 2019) defined in Equation 1. SCE incorporates Reverse Cross Entropy (Equation 2) in addition to the standard Cross Entropy (Equation 3), with tunable hyperparameters and balancing the two components.

$$\ell_{ce} = -\sum_{k=1}^{K} q(k|x) \log(p(k|x)) \quad (2)$$

$$\ell_{rce} = -\sum_{k=1}^{K} p(k|x) \log(q(k|x)) \quad (3)$$

$$\ell_{sce} = \mu \ell_{ce} + \beta \ell_{rce} \quad (4)$$

where $\ell_{ce}$ is the cross-entropy loss $q(k|x)$ is the ground truth class distribution conditioned on sample x $p(k|x)$ is the predicted distribution over labels by the classifier $\ell_{rce}$ is the reverse cross-entropy loss $\ell_{sce}$ is the symmetric cross-entropy loss $\mu$ and $\beta$ are tunable hyperparameters

Other key aspects included oversampling (Appendix D) to handle the imbalanced label distribution and pre-training the RoBERTa model on the back-translated data before fine-tuning on the actual task.

### 5.2 Experiments and Results

We randomly selected 43 essays with perfect scores from zuowenwang[1] as our fluency excellence examples. We utilized Chinese-roberta-wwm-ext[2] as our base model. Based on ablation experiments (Appendix E), we fixed the hyperparameters $\mu$ as 0.1 and $\beta$ as 1 for the Symmetric Cross Entropy (SCE) loss. With this configuration, we achieved the state of the art on the test set with the score of 51.96 for Track 3.

## 6 Conclusions

This report presented our approaches and results for the three tracks of the Chinese Essay Fluency Evaluation (CEFE) task at CCL-2024.

For Track 1, we employed a hierarchical method combining token-level error identification, coarse-grained modeling on the Chinese Learner 4W corpus, fine-grained binary models, and prediction fusion to handle both broad and specific error types.

In Track 2, constructing pseudo-data with multiple error types per sentence improved performance in correcting real-world sentences compared to previous single-error methods.

Our Track 3 approach, which achieved first place, utilized back-translated pseudo-data with triple fluency labels, an NSP-based strategy to incorporate context while mitigating long sequence issues, and Symmetric Cross Entropy loss for increased robustness.

By addressing challenges such as limited data, error diversity, long-range dependencies, and label noise, our methods contribute to advancing intelligent assessment of Chinese essays. Potential future directions include cross-lingual generalization, few-shot learning to reduce annotation requirements, and generating more detailed feedback to further enhance student learning.

## 7 Acknowledgements

This work is partially supported by Guangzhou Science and Technology Plan Project (202201010729), and Guangdong Social Science Foundation Project (GD24CWY11). We thank the reviewers for their helpful comments and suggestions.

---

[1] https://www.zuowen.com/xsczw/fanwen/
[2] https://huggingface.co/hfl/chinese-roberta-wwm-ext

# Appendix A  Track 1: Choice of Training Dataset for Coarse-grained Error Modeling

| Corpus/Strategy | Char F1 | Coll F1 | Miss F1 | Redu F1 |
| --- | --- | --- | --- | --- |
| CSED/repeatable | 0.388 | 0.249 | 0.237 | 0.292 |
| CSED/non-repetitive | 0.401 | 0.302 | 0.311 | 0.320 |
| 4W/repeatable | **0.529** | 0.513 | 0.255 | **0.501** |
| 4W/repetitive | 0.499 | **0.632** | **0.298** | 0.496 |

Table 1: F1 scores for models with different corpora and strategies

Table 1 shows the performance of the two corpora on the four models under different strategies. The experimental analysis shows that the models trained on the 4W corpus constructed by Chinese learners generally outperform those trained on the CSED corpus. When comparing corpus strategies, the uniform allocation strategy performs better in both corpora, while the repetition strategy may lead to overfitting. Therefore, we selected the 4W corpus with the uniform allocation strategy (i.e. **4W/repetitive**) as the best solution for Track1 coarse-grained model training.

**Appendix B    Track 1: Choice of Strategy for Fine-grained Error Categorization**

Table 2 presents the corpus samples, the respective number of sentences constructed using two different corpus construction methods, and the F1 scores obtained from the corresponding trained models.

| Method | Fine-Grained | Label | Sentence Number | F1 |
| --- | --- | --- | --- | --- |
| Wrong - Correct | Misorder | 0 | 50 | 24.5 |
|  |  | 1 | 50 |  |
|  | Redundancy of other constituents | 0 | 194 | 38.4 |
|  |  | 1 | 194 |  |
| Wrong - Variant Error | Misorder | 0 | 50 | 64.9 |
|  |  | 1 | 51 |  |
|  | Redundancy of other constituents | 0 | 194 | 50.4 |
|  |  | 1 | 196 |  |

Table 2: The binary classification model training corpus and the corresponding F1 values for both methods. The ratio of 0 and 1 in both methods is 1:1.

Based on the F1 scores of the two strategies, we decided to train with the second strategy of **Variant Error Sentence**, which involves using a corpus that consists of both specified error types and a variety of other error types.

**Appendix C    Track 2: Validation of Pseudo-data Construction Method**

| Strategy | EM | BLEU | $F_{0.5}$ | B.S. | Leven | $PPL_{BERT}$ | Avgscore |
| --- | --- | --- | --- | --- | --- | --- | --- |
| Bart-base + 1 error | 1.0 | 86.86 | 21.7 | 96.89 | 0.91 | 2.72 | 47.99 |
| Bart-base + 2 errors | 2.0 | 86.64 | 24.35 | 96.92 | 1.16 | 2.67 | 48.65 |
| Bart-base + 3 errors | 3.0 | 86.82 | 25.95 | 96.96 | 1.26 | 2.65 | 49.27 |
| Bart-base + 4 errors | 3.0 | 86.70 | 27.29 | 97.03 | 1.31 | 2.64 | 49.55 |

Table 3: The table presents the results from the pseudo dataset containing numerous errors on the validation set.

The proposed methodology, which involves constructing a pseudo-dataset with sentences containing multiple error types and training an encoder-decoder model on this dataset, proves to be effective in enhancing the performance of the sentence rewriting model. The experimental results (Table 3) demonstrate the superiority of this approach compared to training on sentences with only a single error type.

**Appendix D    Track 3: Sampling Strategy for Back-translation**

We observe that the distribution of multi-label quantities is imbalanced (Table on the left of Figure 5), for which we adopt the oversampling strategy (figure on the right of Figure 5).

| Label | Nums | Percent. |
|---|---|---|
| Excellent (优秀) | 12 | 12% |
| Moderate (一般) | 45 | 45% |
| Failing (不及格) | 43 | 43% |

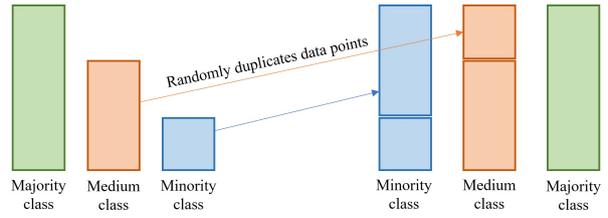

Figure 5: The table on the left displays the distribution of the multi-label quantities. And the figure on the right illustrates the oversampling strategy.

## Appendix E  Track 3: Choice of Parameters for SCE Loss

Table 4 presents the ablation studies evaluating different $\mu$ and $\beta$ parameter values for the SCE loss:

| $\mu$ | $\beta$ | Acc | F1 | QWK | AvgScore |
|---|---|---|---|---|---|
| 1 | 0 | 50.0 | 36.67 | 62.26 | 45.79 |
| 1 | 1 | 50.0 | 37.23 | 69.51 | 47.52 |
| 0.5 | 1 | 60.0 | 64.44 | 79.59 | 66.14 |
| 0.1 | 1 | **70.0** | **74.6** | **85.85** | **75.47** |
| 0.05 | 1 | 60.0 | 64.44 | 79.59 | 66.14 |
| 0.01 | 1 | **70.0** | 65.56 | 71.7 | 68.12 |

Table 4: The ablation studies for searching the optimal parameters in validation. Notice that: when $\beta$ is set to 0, cross-entropy loss is employed during training.

We decided that $\mu = 0.1$ and $\beta = 1$ provided the best balance between standard and reverse cross-entropy for robust training on the potentially noisy pseudo-data labels.